\documentclass[letterpaper]{article}
\usepackage{aaai2026} 
\usepackage{times} 
\usepackage{helvet} 
\usepackage{courier} 
\usepackage[hyphens]{url} 
\usepackage{graphicx} 
\usepackage{graphicx} 
\usepackage{natbib} 
\usepackage{caption} 
\usepackage{amsfonts}
\usepackage{amsmath}
\DeclareMathOperator*{\argmin}{arg\,min}
\usepackage{booktabs}
\usepackage{tabularx}
\usepackage{todonotes}
\usepackage{array}

\title{Quiet Feature Learning in Algorithmic Tasks}
\author{
    Prudhviraj Naidu, Zixian Wang, Leon Bergen\equalcontrib, Ramamohan Paturi\equalcontrib
}
\affiliations{
    Department of Computer Science\\
    UC San Diego\\
    San Diego, CA 92093, USA \\
    prnaidu@ucsd.edu,
    ziw081@ucsd.edu,
    lbergen@ucsd.edu,
    rpaturi@ucsd.edu
}

\begin{document}

\maketitle

\begin{abstract}
We train Transformer-based language models on ten foundational algorithmic tasks and observe pronounced \emph{phase transitions} in their loss curves that deviate from established power-law scaling trends. Over large ranges of compute, the validation loss barely improves, then abruptly decreases. Probing the models’ internal representations reveals that \textit{quiet features} are learned prior to any decrease in task loss. These quiet features represent intermediate algorithmic computations that do not by themselves improve the output loss. Ablation experiments demonstrate that individual quiet features are causally necessary for task performance. Our results demonstrate that substantial representational progress can remain hidden beneath an apparently flat loss curve, challenging the prevailing use of cross‑entropy as a proxy for learning and motivating richer diagnostics for monitoring model training.
\end{abstract}

\begin{links}
    \link{Code}{https://github.com/prudhvirajn/quiet-feature-learning-in-algorithmic-tasks}
\end{links}

\section{Introduction}

Understanding how and when large language models acquire new capabilities has become an important question in deep learning. While language models demonstrated remarkable performance across a broad range of tasks, the precise mechanisms driving their improvements remain unknown. Recent discussions of ``emergent abilities" -- where larger-scale models outperform baselines abruptly, even though smaller-scale counterparts exhibit little improvement -- have led to debate over whether such phenomena are genuine or artifacts of measurement \citep{wei2022emergent,GanguliHLABCCDD22,SchaefferMK23}.

Questions about emergent abilities are closely tied to the observation of scaling laws in model training \citep{kaplan2020scaling, ruan2024scalingformodelfaimilies, henighan2020scalinglawsautoregressivegenerative, llama32024herdofmodels, gpt4technicalreport2023}. These scaling laws typically show a smooth, power-law relationship between compute and model performance. However, most empirical demonstrations of these laws derive from heterogeneous data and tasks, leaving open the possibility that ``averaging out" many distinct learning behaviors masks more abrupt transitions that occur for individual skills or subtasks.

In order to better understand skill learning in a tractable setting, we focus on ten foundational algorithmic problems spanning various input types. These algorithmic tasks have precisely defined solutions, making it straightforward to identify clear success criteria, isolate the specific features the model must learn, and ensure that improvements cannot be attributed to memorization or partial heuristics. These tasks allow us to investigate fine-grained learning phenomena which might otherwise be obscured by heterogeneous data.

Our key findings include:
\begin{enumerate}
    \item \textbf{Phase transitions occur during learning}: We observe two distinct phases in scaling laws across tasks and input sizes. In the \textit{slow phase}, loss improves minimally or remains flat. Then, loss drops rapidly (\textit{fast phase}). We refer to the change between these two phases as a phase transition. Phase transitions occur for scaling laws estimated across many training runs and within individual training runs. 
    
    \item \textbf{\textit{Quiet features} precede phase transitions}: Models learn meaningful internal representations during the slow phase, but these features do not yet yield noticeable gains in the output loss (we call these \textit{quiet features}). Ablating them severely degrades performance, demonstrating they are causally related to the eventual sharp drop in loss.

\end{enumerate}

These findings challenge the assumption that improvements in loss necessarily coincide with improvements in feature representations.
Instead, substantive internal reorganization may occur below the surface, revealing itself only at discrete points during training.

The rest of this paper is organized as follows: Section \ref{section:related-work} reviews related work in scaling laws, emergent abilities, and algorithmic learning. Section \ref{section:phase-changes} describes our experimental methodology and presents our observations of phase transitions across tasks and input sizes. Section \ref{section:feature-learning} introduces our feature analysis framework and demonstrates how quiet and loud features evolve during training. Finally, Section \ref{section:discussion} discusses the broader implications of our findings and suggests directions for future research.

\begin{figure*}[ht]
\includegraphics[width=\linewidth]{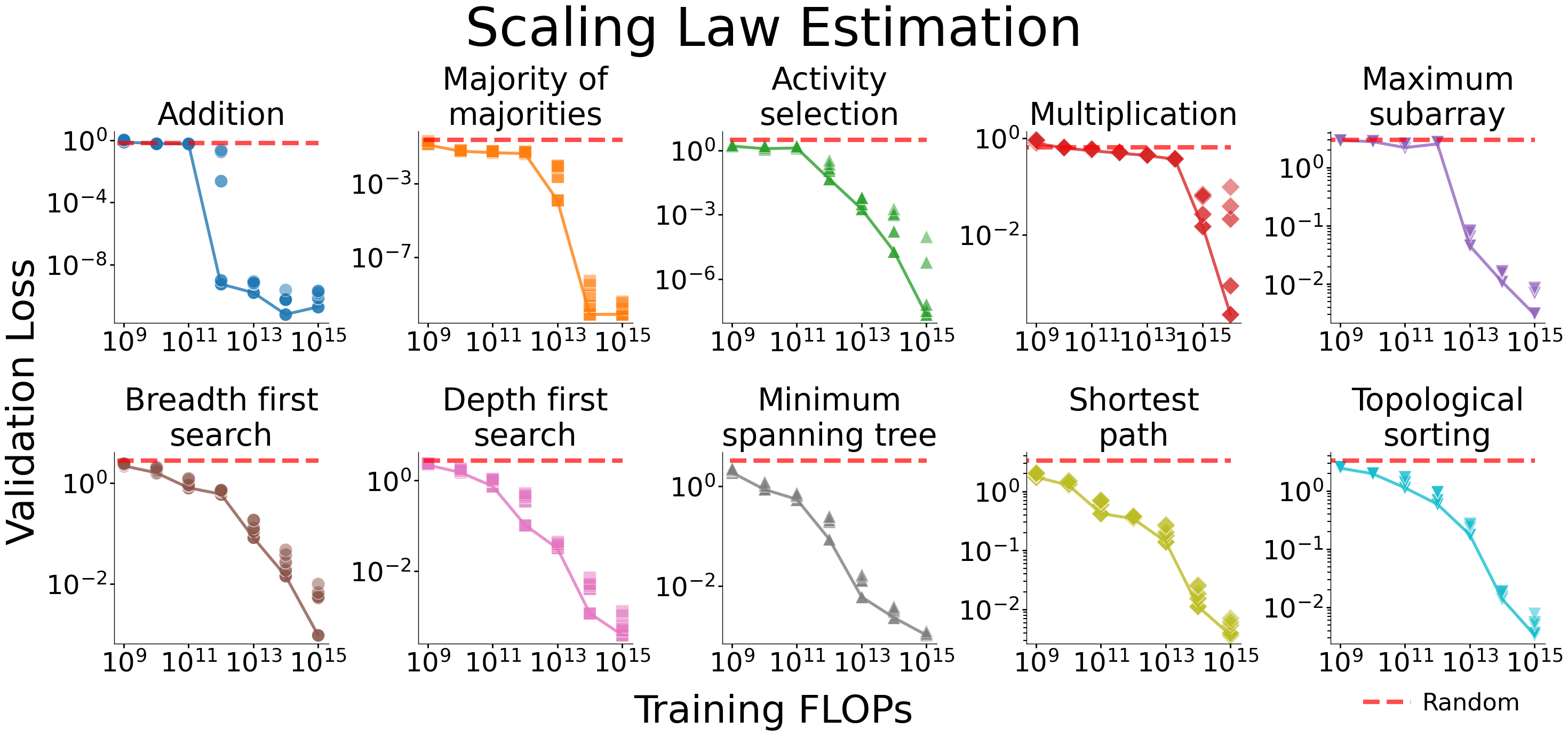}
\caption{Model performance (validation loss) abruptly improves as we increase the model size, dataset size, and amount of compute (Training FLOPs) used for training. The input size for addition, multiplication and activity selection is 16. For graph tasks, the input size is 11. For maximum subarray, the input size is 64 while for majority of majorities it is 32. The red dotted line indicates random performance.}
\label{fig:val-loss-across-tasks}
\end{figure*}

\section{Related Work}
\label{section:related-work}

\subsection{Scaling Laws}

\citet{hestness2017deeplearningscalingpredictable} observed scaling dataset size and deep neural network model size led to a predictable decrease in generalization error for neural machine translation, language modeling, image classification and speech recognition. \citet{kaplan2020scaling} and \citet{hoffman2022} observed predictable relationships between training compute and language modeling loss. 
\citet{henighan2020scalinglawsautoregressivegenerative} extended this work for generative models across modalities: image, video, multimedia image-text and math. They demonstrated classification loss and error rates predictably decreased on downstream image classification tasks. \citet{chentworek2021} studied language model performance on coding. They observed a predictable relationship between language modeling loss on a held out code corpus and model size.

\subsection{Predicting LLM abilities}

\citet{GanguliHLABCCDD22} and \citet{wei2022emergent} demonstrated some large language models' capabilities could not be predicted from capabilities of small language models. However, \citet{SchaefferMK23}, \citet{gpt4technicalreport2023}, \citet{ruan2024scalingformodelfaimilies}, and \citet{llama32024herdofmodels} provide evidence that this is due to choice of metrics and that large language models capabilities can be predicted from small language models.

\subsection{Proposed explanations for scaling laws}

\citet{MichaudLGT23} proposed neural networks learn discrete skills called ``quanta." They argue that there is a strict ordering, which they called Q sequence, in which quanta must be learned, and that the frequencies of these quanta follow a power law, leading to the power law relationship observed by \citet{kaplan2020scaling} and others. \citet{marcushutter2021} propose that the relationship between the error rate and dataset size is guided by the distribution of features in the data. They show a Zipfian distribution of features results in power law scaling. 


\subsection{Grokking}
In grokking \citep{power2022grokking, NandaCLSS23, Varma2023}, a model trained for many epochs quickly memorizes the training set (thus achieving high training accuracy early) but only later learns a generalizing solution, causing a sudden jump in test accuracy. Our scaling law results are related to grokking, but occur in the single epoch setting. Unlike grokking, models trained in the single-epoch setting do not exhibit a transition from memorization to generalization.

\begin{figure*}[ht]
\centering
\includegraphics[width=\linewidth]{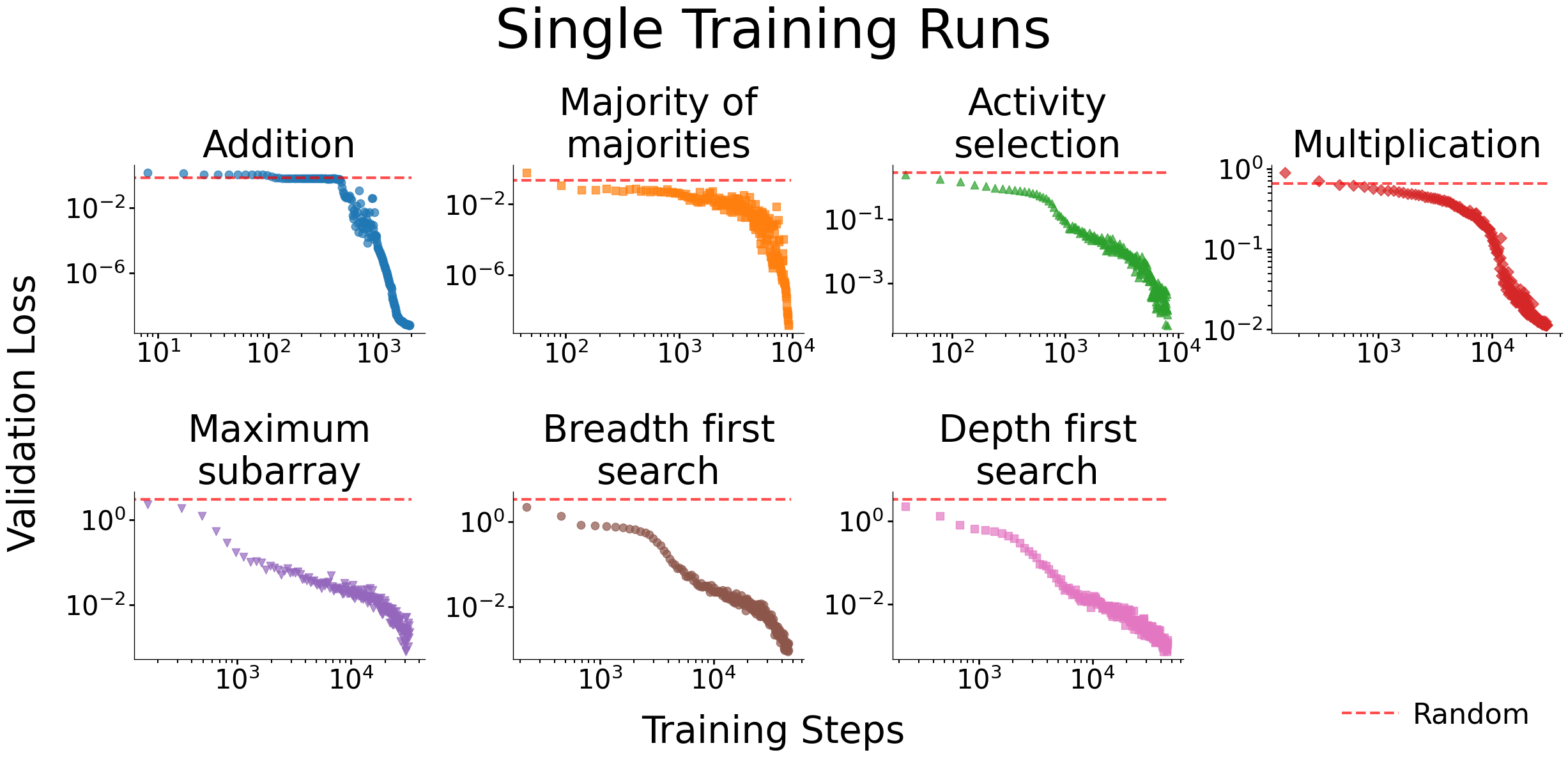}
\caption{Model exhibit similar abrupt improvement in performance during a single training run. Plots show compute-optimal training runs for the smallest compute budget where test accuracy is 100\%. The red dotted line indicates random performance.}
\label{fig:val-loss-training-steps-multiple-tasks}
\end{figure*}

\subsection{Progress Measures}
Several previous works have identified measures which track progress toward the final, fully-generalizing solution, even when the test loss shows no improvement. 
\citet{BarakEGKMZ22} propose a metric for measuring similarity of network weights in the context of sparse parity, and demonstrate that this metric continuously improves throughout training, including prior to measurable improvement in generalization performance. \citet{NandaCLSS23} propose a different metric on network weights in the context of modular arithmetic grokking, and demonstrate that this metric improves before the phase transition. \citet{RFMGrokking2024} propose tracking features using Average Gradient Outer Product (AGOP) for Recursive Feature Machines. While this prior work has focused on measuring progress in model weights, they do not demonstrate that the networks are computing interpretable activations prior to generalization. We close this gap by directly probing for human-interpretable features and showing they appear well before the loss drop.

\subsection{Phase Transitions}

Phase transitions were previously observed for a limited number of algorithmic tasks. \citet{olsson2022context}, \citet{GDPG2022}, and \citet{ETEMG2024} find phase transitions for in-context learning during individual training runs. \citet{BarakEGKMZ22} observed phase transitions in parity. \citet{LeeSL0P24} measures relationships between test accuracy and number of examples (over a fixed model size), with observed phase transitions potentially being explained by the metric artifacts of \citet{SchaefferMK23}. 

\section{Scaling Laws for Algorithmic Tasks}
We first aim to estimate scaling laws for 10 foundational algorithmic tasks. Scaling laws are estimated by training models over a range of compute budgets, and identifying the optimal model at each budget.
\label{section:phase-changes}

\subsection{Task Formulation}
 
We examine 10 algorithmic tasks which are drawn from three broad categories: binary arithmetic, graph algorithms and sequence-based optimization. The tasks capture a range of input types, and have well-understood algorithms for solving them.

All tasks are formulated as sequence prediction problems. The input to the problem is serialized, and an autoregressive model is trained to predict the solution. All tasks use a standard cross-entropy loss, with the loss masked on the input tokens. We describe how we formulate three of the tasks below. For other tasks, please see Technical Appendix A. 

\subsubsection{Binary Addition}
We formulate $n$-bit binary addition as the following sequence prediction task:
$$\texttt{x}_1\texttt{x}_2\ldots\texttt{x}_n\texttt{+}\texttt{y}_1\texttt{y}_2\ldots\texttt{y}_n\texttt{=}\texttt{z}_1\texttt{z}_2\ldots\texttt{z}_{n+1}\texttt{<EOS>}$$
where $\texttt{x}$, $\texttt{y}$, and $\texttt{z}$ are binary numbers, presented from the least significant bit to the most significant bit. Each bit is represented as a separate token, and $\texttt{+}$, $\texttt{=}$, and $\texttt{<EOS>}$ are also represented as individual tokens. 

\subsubsection{Breadth First Search}

Given a connected undirected graph $G$ with $n$ vertices $V = \{v_1, v_2, \ldots, v_n\}$, a set of edges $E$, and a start vertex $v_s$, the task is to predict the traversal order in a breadth first search.

We formulate this as:
$$\texttt{v}_s\texttt{v}_{i_1}\texttt{v}_{j_1}\ldots\texttt{v}_{i_m}\texttt{v}_{j_m}\texttt{=}\texttt{v}_{t_1}\texttt{v}_{t_2}\ldots\texttt{v}_{t_n}\texttt{<EOS>}$$
where $(\texttt{v}_{i_k}, \texttt{v}_{j_k})$ represents an edge in $E$, $m = |E|$ is the number of edges, and $\texttt{v}_{t_1}\texttt{v}_{t_2}\ldots\texttt{v}_{t_n}$ is the complete BFS traversal sequence starting from $v_s$ (where $\texttt{v}_{t_1} = \texttt{v}_s$). Ties in BFS ordering are broken by lexicographic ordering.

\subsubsection{Maximum Subarray}
Given a sequence of $n$ integers $k_1, k_2, \ldots, k_n$ where $k_i \in [-9, 9]$, the maximum subarray task is to predict the contiguous subarray with the maximum sum.

We formulate this as:
$$\texttt{k}_1\texttt{k}_2\ldots\texttt{k}_n\texttt{=}\texttt{k}_i\texttt{k}_{i+1}\ldots\texttt{k}_j\texttt{<EOS>}$$
Where $\texttt{k}_i\texttt{k}_{i+1}\ldots\texttt{k}_j$ is the maximum sum subarray ($i \leq j$). 

\subsection{Experimental Methodology}

\begin{figure*}[ht]
\includegraphics[width=\linewidth]{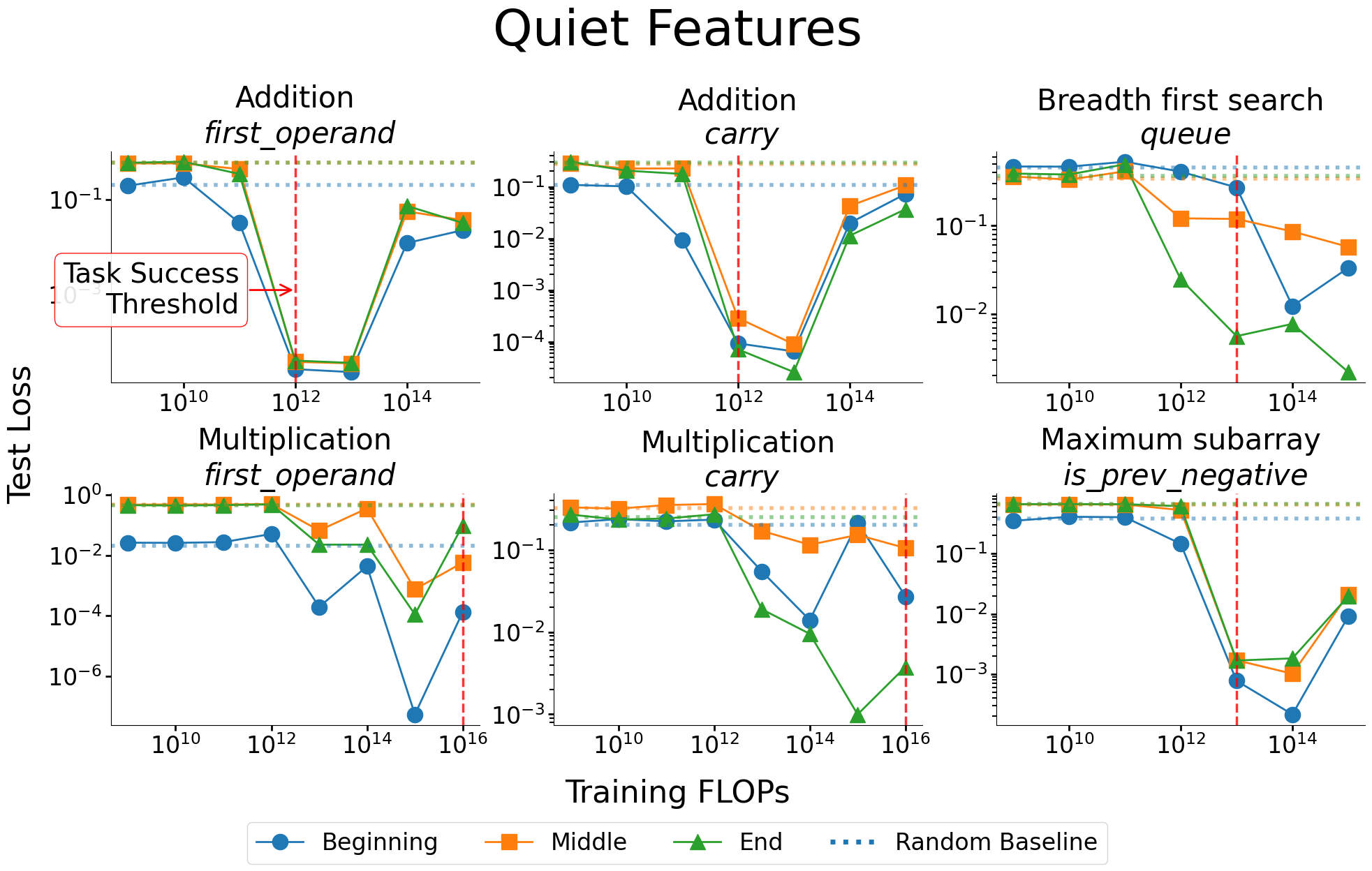}
\caption{Models learn quiet features before the phase transition. The loss is averaged over the first third of token positions (Beginning), second third (Middle), and last third (End). The red vertical line indicates the task success threshold, which is the smallest compute budget at which the task loss starts to decrease (see Technical Appendix Figure 6). Horizontal dotted lines represent random baselines.}
\label{fig:quiet-features-before-phase-change}
\end{figure*}


\subsubsection{Model Training}

Each task is trained independently with the Transformer++ architecture. Transformer++ is a decoder-only transformer model with enhancements detailed in Technical Appendix Table 4, based on modifications in Llama and PALM \citep{gu2024mamba}. This architecture is chosen because it has improved performance in scaling law experiments compared to other transformer variants \citep{gu2024mamba}.


Models are trained with the AdamW optimizer \citep{loshchilov2017} with linear warmup followed with cosine learning rate annealing as prescribed by \citet{hoffman2022}.

\subsubsection{Estimating Scaling Laws}
The scaling law experiments aim to estimate the best performance achievable on a task given a compute budget. Separate scaling laws are estimated for each task and input size. Each model is trained up to a pre-specified compute budget, which ranged from $10^9-10^{15}$ FLOPs.\footnote{For multiplication, the maximum budget was increased to $10^{16}$, since this was the minimum budget needed to train the task to 100\% accuracy.} For each budget, we conduct a grid search across model sizes, batch sizes, and learning rates (see Technical Appendix Table 3 for details about the hyperparameter search). Following the procedure from Chinchilla \citep{hoffman2022}, the period of the learning rate scheduler is set to the number of training steps.



The number of training runs per task varies from 1316 to 3565, and the total number of training runs is 18544. All models are trained for at most a single epoch; each algorithmic task has a sufficient number of unique examples to avoid repetition even with the highest compute budgets. The number of training examples is determined based on training compute budget and model size, with all configurations evaluated using randomly generated validation and test sets with 1000 examples each. We choose the configuration with minimum validation loss for each training compute and designate it as the ``compute-optimal validation loss."

\begin{figure*}[ht]
\centering
\includegraphics[width=0.95\linewidth]{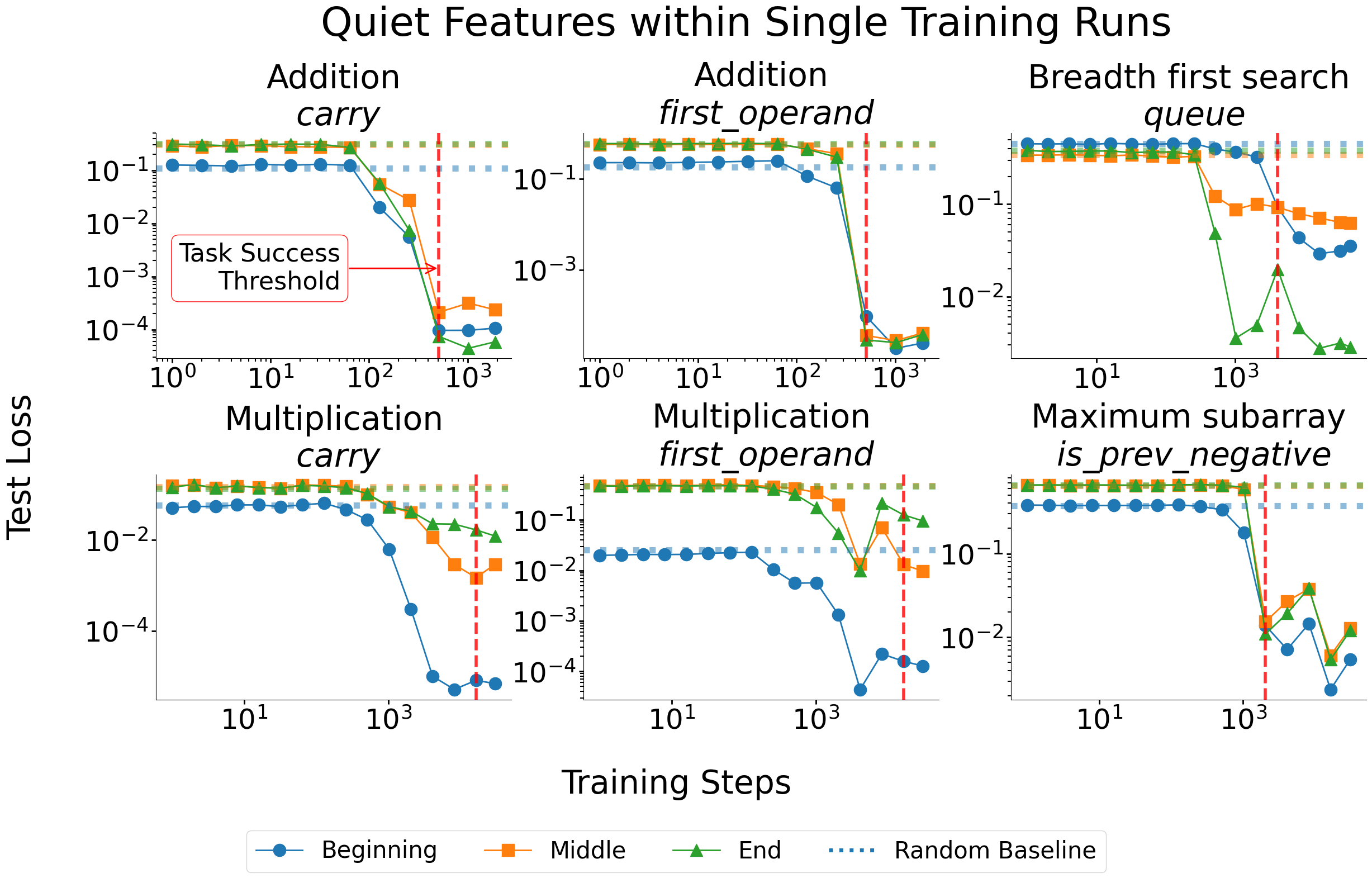}
\caption{Models learn quiet features before the phase transition within single training runs. The loss is averaged over the first third of token positions
(Beginning), second third (Middle), and last third (End). Plots show compute-optimal training runs for the smallest compute budget where test accuracy is 100\%. The red vertical line indicates the task success threshold, which is the training step at which the task loss starts to decrease. Horizontal dotted lines represent random baselines.}
\label{fig:quiet-features-during-training-run}
\end{figure*}

\subsection{Scaling Law Results}
\label{subsec:scaling-results}

We observe phase transitions for compute-optimal validation loss across three scenarios: (1) when we vary both model size \& dataset size, (2) when we fix the model size \& vary the dataset size, and (3) during individual (compute-optimal) training runs. Figure \ref{fig:val-loss-across-tasks} shows that for six of the tasks, the compute-optimal validation loss undergoes a clear phase transition as the training compute budget increases. For these tasks, there are two distinct phases of learning: a slow phase and a fast phase. During the slow phase, loss is stagnant or decreasing slowly. During the fast phase, the loss decreases rapidly.

For addition, majority of majorities, activity selection and maximum subarray the validation loss is roughly constant in the slow phase then suddenly goes to near zero during the fast phase. For multiplication and breadth first search, the slow phase has a gradual decrease followed by a steeper decrease in the fast phase.

Next, we investigate the effects of varying the dataset size. In Technical Appendix Figure 6, we fix the model size (selecting the model size corresponding to the smallest training compute budget that achieves 100\% test accuracy) and increase the dataset size. We continue to observe phase transitions even when the model size is fixed. In this setting, additional graph tasks exhibit distinct phase transitions. 



We next analyze model behavior within individual training runs.\footnote{These training runs correspond to compute-optimal  hyperparameter settings.} Figure \ref{fig:val-loss-training-steps-multiple-tasks} shows these individual training runs exhibit phase transitions in the loss. For addition and majority of majorities, there is a predictable power-law regime after the phase transition.


Phase transitions in compute-optimal validation losses occur across different task sizes (see Technical Appendix Figure 8). For addition, phase transitions are observed across task sizes and similarly within individual training runs (Technical Appendix Figure 9). As the input size increases, the Pareto frontier shifts to the right but maintains the same shape. However, for maximum subarray, the phase transition only appears at task sizes greater than 16.



\begin{figure*}[ht]
\centering
\includegraphics[width=\linewidth]{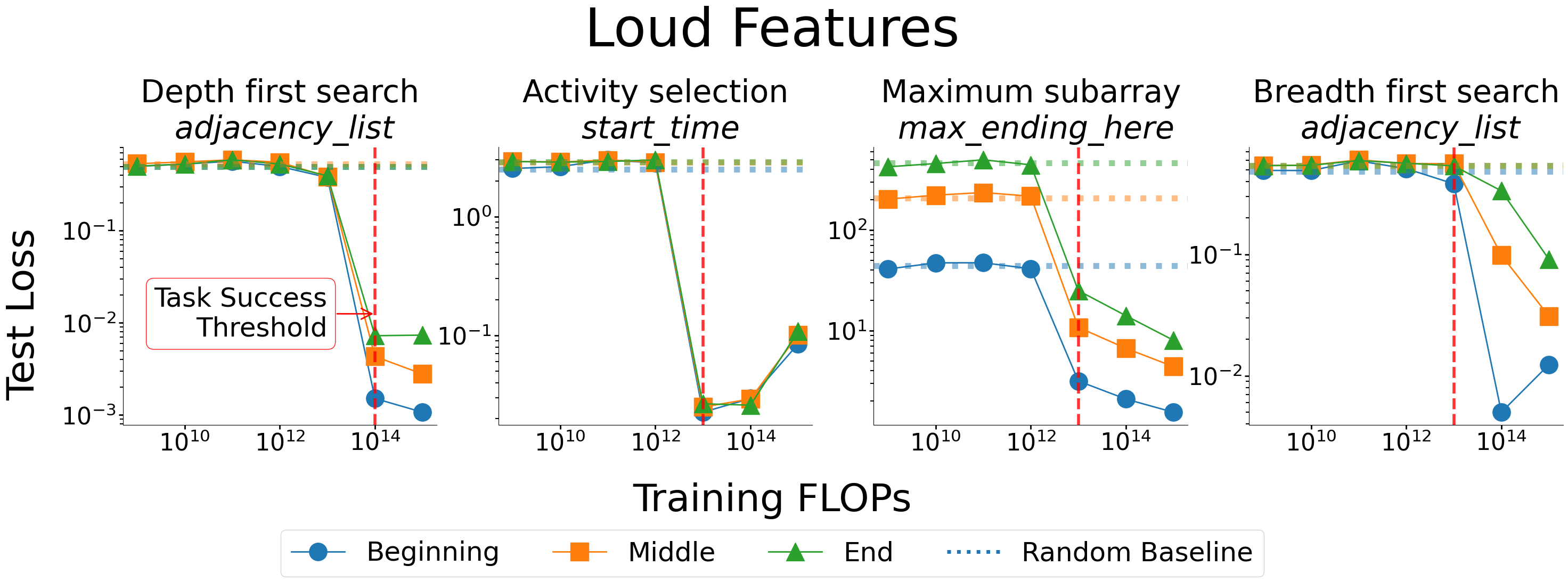}
\caption{Models learn different set of features (loud features) at or after the phase transition. The loss is averaged over the first third of token positions (Beginning), second third (Middle), and last third (End). The red vertical line indicates the task success
threshold, which is the smallest compute budget at which the task loss starts to decrease (see Technical Appendix Figure 6). Cross-entropy loss is used for training probes for \textit{first\_operand}, \textit{adjacency\_list}. The probing loss is mean squared error for \textit{start\_time} and \textit{max\_ending\_here}, since these are continuous features.}
\label{fig:feature-all-during-phase-change}
\end{figure*}


\section{Feature Learning before Phase Transitions}
\label{section:feature-learning}

In order to better understand the observed phase transitions, we investigate the emergence of human-interpretable features during learning. We focus on features corresponding to intermediate outputs of standard algorithms used to perform the tasks. We use linear probing to identify whether the model learned these features. 

\subsection{Feature Probing Methodology}
\label{subsection:probing-method}
For each algorithm-specific feature, we train separate linear probes across each token position and each layer. Probes are trained on the residual streams after each layer (See Technical Appendix B). Each probe is trained with 10,000 examples which had been held-out from the original model training set. 


For each task, we aim to identify the smallest compute budget at which a feature emerges. We select a single model size to study for this task; models of this size are trained for different compute budgets.\footnote{The model size is chosen so that it is nearly optimal across compute budgets. A fixed model size is chosen in order to make feature metrics easier to compare for models across training runs.} We train separate linear probes for each (model, token position, layer) triple. For each (model, token position) pair, we select the probe that achieves the lowest training loss across layers. We report the test performance of the selected probe for each (model, token position) pair.

We establish random baselines by applying the same probing methodology to models initialized with random weights. Test loss is estimated on a separate test set of 1,000 unseen examples. 

\subsection{Intermediate Task Features}
We describe the features investigated for each of the tasks which exhibit phase transitions in their loss. These features are intermediate values computed in standard algorithms for the tasks.
\paragraph{Addition \& Multiplication.} For n-bit binary addition, we probe for the following at each token $\texttt{z}_i$: \textit{first\_operand}, which is input bit $\texttt{x}_{i+1}$ (required to compute $\texttt{z}_{i+1}$); and \textit{carry} $c_i$, the carry bit used to compute $\texttt{z}_{i+1}$. Carry $c_0$ for $z_1$ is not considered since the first carry is always zero. For multiplication, we check whether the model learns carries generated when adding the last partial product to the sum of the previous $n-1$ partial products. 

\paragraph{Breadth/Depth First Search.} For breadth first search, we probe at each token $\texttt{v}_{t_i}$ for the following: \textit{queue}, which is the set of vertices on the queue (in the standard search algorithms) after we have explored vertex $\texttt{v}_{t_i}$; and \textit{adjacency\_list}, which is the set of vertices adjacent to $\texttt{v}_{t_i}$.

\paragraph{Maximum Subarray.} For the maximum subarray problem, we probe at each token $\texttt{k}_{i}$ (before the \texttt{=} token) for: \textit{is\_prev\_negative}, which represents whether $\texttt{k}_{i-1}$ is negative; and $\textit{max\_ending\_here}$, which is the maximum sum of the contiguous subarray ending at $\texttt{k}_{i}$. (Refer to \citet{Kadane23} for the standard algorithm.)

\paragraph{Activity Selection.} For the activity selection problem, we probe at each token $\texttt{f}_{i}$ for \textit{start\_time}, which is the corresponding start time $\texttt{s}_i$. Since the model has to output $\texttt{s}_i\texttt{f}_i$ in order, it must know which start times correspond to which finish times. (Refer to \citet{kleinberg2005} for the standard algorithm.)

\begin{table}[ht]
\begin{tabularx}{\linewidth}{lX>{\raggedleft\arraybackslash}X}
\toprule
\textbf{Task} & \textbf{Feature} & \textbf{Feature Ablation} \\ 
& & $\Delta$ Accuracy (\%) \\
\midrule
Addition (16) & \textit{carry} & $-41.2^{*}$ \\ 
Addition (32) & \textit{carry} & $-50.4^{*}$ \\
Addition (64) & \textit{carry} & $-75.1^{*}$ \\ 
Addition (16) & \textit{first\_operand} & $0.00$  \\ 
Addition (32) & \textit{first\_operand} & $-92.7^{*}$ \\ 
Addition (64) & \textit{first\_operand} & $-6.40^{*}$ \\
Multiplication (16) & \textit{carry} & $-20.3^{*}$ \\ 
Multiplication (16) & \textit{first\_operand} & $-0.05$ \\ 
Maximum Subarray (64) & \textit{is\_prev\_negative} & $-4.14^{*}$ \\ 
Breadth first search (11) & \textit{queue} & $-43.6^{*}$ \\
\bottomrule
\end{tabularx}
\caption{Average difference in test accuracy after ablating a quiet feature compared to ablating a random direction (random ablation). Ablating quiet features degrades test accuracy more than random ablation. For random ablation, we estimate test accuracy over 32 trials. $^{*}$ indicates $ p < 0.001$ using bootstrapping. For complete accuracy / loss values see Technical Appendix Table 6} 
\label{tab:ablation}
\end{table}


\subsection{Feature Probing Results}


The model learns algorithmic features before, during and after the phase transition. We call features learned prior to the phase transition \textit{quiet features}, as they occur during the slow phase where loss is stagnant or slowly decreasing. Features learned in the fast phase (during and after the phase transition) are \textit{loud features} as the task loss decreases rapidly. Figure \ref{fig:quiet-features-before-phase-change} shows the trajectory of the probing loss for \textit{quiet features}. For addition, multiplication, and maximum subarray, the model learns features for early token positions prior to the phase transition. However, for breadth-first search, later token positions are learned first. 

These results apply across distinct, compute-optimal training runs (for a fixed model size). Figure \ref{fig:quiet-features-during-training-run} shows that quiet features also emerge during individual training runs. Features for early token positions are also generally learned first in this case.

Figure \ref{fig:feature-all-during-phase-change} shows models learns \textit{loud features} in the fast phase (during and after the phase transition).

A surprising finding is the U-shape of feature learning curves in Figure \ref{fig:quiet-features-before-phase-change}, indicating that the probing loss increases for many quiet features after the phase transition. This may indicate that the models are learning alternative representations in the highest compute budget regimes, though the probing loss remains below the random baseline.

\subsection{Are \textit{quiet features} causal?}

For a given task, we ablate a \textit{quiet feature} from the residual stream at each position, using the feature probes (one probe per position) identified the previous section. We restrict our analyses to binary features. By comparing to ablations of random features, we can evaluate whether a quiet feature is causally responsible for task performance.

We ablate a feature by removing its direction from the residual stream. A linear feature probe $w^\top x^* + b$ outputs $0$ (assigns $0.5$ probability to each label) when it detects no information from the ablated residual stream at that layer. Letting $x$ be the residual stream at a desired layer, we perform the following optimization:
\[
\begin{aligned}
\operatorname*{argmin}_{x^*}\quad & \|x - x^*\|^2 \\
\text{subject to}\quad & w^\top x^* + b = 0.
\end{aligned}
\]
Solving this yields $x^* = x - \frac{w^\top x + b}{||w||^2} w$. The residual stream activation at the linear probe's layer is replaced with $x^*$.

Ablation results are shown in Table \ref{tab:ablation}. Quiet feature ablations are compared to ablating a random direction. When we ablate quiet features, we observe test accuracy generally degrades more than ablating a random direction, indicating a causal role for quiet features. Similar results are seen for test loss, as shown in Technical Appendix Table 6. However, for \textit{first\_operand} at input size 16, we do not see any significant change after ablating the feature compared to random in addition or multiplication. At larger input sizes, ablating  \textit{first\_operand} for addition leads to significant test loss degradation compared to random ablation.



\section{Discussion}
\label{section:discussion}

Our findings show that, across different algorithmic tasks, there is often a long phase of training with little apparent improvement in next-token prediction loss. Despite this plateau, we observe that essential internal features (e.g., carry bits in binary addition, adjacency in breadth first search) emerge during these periods. These quiet features emerge prior to any substantial improvement in task performance. Ablation experiments confirm that these features are causally important to solving the tasks, suggesting that models can accumulate partial competence that does not immediately translate into lower loss.

One reason for this quiet period may be the all-or-nothing nature of these tasks: obtaining just some of the required subroutines (e.g., some correct carry bits) does not prevent errors on next token prediction. Consequently, any reduction in loss is small until all sub-features are aligned. In over-parameterized models, there is sufficient capacity to learn these subroutines in the background, allowing partial solutions to remain in the representations until they can be combined into a correct overall procedure.

These findings have practical and conceptual implications. For practitioners, they highlight the risk of judging model capabilities based solely on loss curves. Probe or circuit-based diagnostics could provide earlier warnings that a model is nearing a capability threshold. Conceptually, they raise questions about whether similar quiet phases exist in more complex natural‑language settings. They also underscore the need for theoretical frameworks that explain why models accumulate latent subroutines before they begin to pay off in observable metrics.

\section{Conclusion}

We observe Transformer-based models for algorithmic tasks encode important intermediate computations well before they show significant gains in next-token prediction. This \textit{quiet period} exposes a gap between internal representation learning and external task performance, indicating that sub-features may lie dormant until the final pieces align. We hope these insights motivate new methods for probing and monitoring internal learning dynamics -- particularly in larger, more complex models -- where hidden phases of progress may likewise precede sudden improvements in capability.

\bibliography{aaai2026}

\clearpage

\appendix

\twocolumn[
\begin{center}
\Large \textbf{Technical Appendix}
\end{center}
]
\section{Task formulation}
\label{sec:task-formulation}

\textbf{Binary Addition}

Binary addition involves adding two $n$-bit numbers to produce an $(n+1)$-bit result. We formulate this as a sequence prediction task:

$$\texttt{x}_1\texttt{x}_2\ldots\texttt{x}_n\texttt{+}\texttt{y}_1\texttt{y}_2\ldots\texttt{y}_n\texttt{=}\texttt{z}_1\texttt{z}_2\ldots\texttt{z}_{n+1}\texttt{<EOS>}$$

Where $\texttt{x}$, $\texttt{y}$, and $\texttt{z}$ represent binary numbers, with $\texttt{x}_1$ denoting the least significant bit (LSB). Each bit is represented as a separate token, and $\texttt{+}$, $\texttt{=}$, and $\texttt{<EOS>}$ are special tokens.

\textbf{Binary Multiplication}

Binary multiplication combines two $n$-bit numbers to produce a $2n$-bit result. We formulate this as:

$$\texttt{x}_1\texttt{x}_2\ldots\texttt{x}_n\texttt{*}\texttt{y}_1\texttt{y}_2\ldots\texttt{y}_n\texttt{=}\texttt{z}_1\texttt{z}_2\ldots\texttt{z}_{2n}\texttt{<EOS>}$$

Following the same convention as in binary addition, with bits ordered from least significant bit to most significant bit.

\textbf{Majority of Majorities}

Given an $n$-bit number (where $n$ is divisible by 4), we partition the bits into 4 equal consecutive groups. For each group, we compute its majority bit value $g_i$. The final output is the majority bit value among $g_1, g_2, g_3, g_4$.

We formulate this as:

$$\texttt{x}_1\texttt{x}_2\ldots\texttt{x}_n\texttt{=}\texttt{z}_1\texttt{<EOS>}$$

Where $\texttt{z}_1$ is the final majority bit.

\textbf{Breadth First Search}

Given a connected undirected graph $G$ with $n$ vertices $V = \{v_1, v_2, \ldots, v_n\}$, a set of edges $E$, and a start vertex $v_s$, we predict the BFS traversal order.

We formulate this as:

$$\texttt{v}_s\texttt{v}_{i_1}\texttt{v}_{j_1}\ldots\texttt{v}_{i_m}\texttt{v}_{j_m}\texttt{=}\texttt{v}_{t_1}\texttt{v}_{t_2}\ldots\texttt{v}_{t_n}\texttt{<EOS>}$$

Where $(\texttt{v}_{i_k}, \texttt{v}_{j_k})$ represents an edge in $E$, $m = |E|$ is the number of edges, and $\texttt{v}_{t_1}\texttt{v}_{t_2}\ldots\texttt{v}_{t_n}$ is the complete BFS traversal sequence starting from $v_s$ (where $\texttt{v}_{t_1} = \texttt{v}_s$).

\textbf{Depth First Search}

This follows the same formulation as BFS, but the expected output $\texttt{v}_{t_1}\texttt{v}_{t_2}\ldots\texttt{v}_{t_n}$ represents the DFS traversal order:

$$\texttt{v}_s\texttt{v}_{i_1}\texttt{v}_{j_1}\ldots\texttt{v}_{i_m}\texttt{v}_{j_m}\texttt{=}\texttt{v}_{t_1}\texttt{v}_{t_2}\ldots\texttt{v}_{t_n}\texttt{<EOS>}$$

\textbf{Shortest Path}

Given a connected undirected graph $G$ with $n$ vertices, a set of edges $E$, and two vertices $v_s$ (source) and $v_f$ (destination), we predict the shortest path between them.

We formulate this as:

$$\texttt{v}_s\texttt{v}_f\texttt{v}_{i_1}\texttt{v}_{j_1}\ldots\texttt{v}_{i_m}\texttt{v}_{j_m}\texttt{=}\texttt{v}_{p_1}\texttt{v}_{p_2}\ldots\texttt{v}_{p_k}\texttt{<EOS>}$$

Where $\texttt{v}_{p_1}\texttt{v}_{p_2}\ldots\texttt{v}_{p_k}$ is the shortest path from $v_s$ to $v_f$ (with $\texttt{v}_{p_1} = \texttt{v}_s$ and $\texttt{v}_{p_k} = \texttt{v}_f$).

\textbf{Topological Sorting}

Given a directed acyclic graph (DAG) $G$ with $n$ vertices and a set of edges $E$, we predict a valid topological ordering of vertices.

We formulate this as:

$$\texttt{v}_{i_1}\texttt{v}_{j_1}\ldots\texttt{v}_{i_m}\texttt{v}_{j_m}\texttt{=}\texttt{v}_{t_1}\texttt{v}_{t_2}\ldots\texttt{v}_{t_n}\texttt{<EOS>}$$

Where $(\texttt{v}_{i_k}, \texttt{v}_{j_k})$ represents a directed edge from $\texttt{v}_{i_k}$ to $\texttt{v}_{j_k}$, and $\texttt{v}_{t_1}\texttt{v}_{t_2}\ldots\texttt{v}_{t_n}$ is a valid topological ordering.

\textbf{Minimum Spanning Tree}

Given a connected undirected graph $G$ with $n$ vertices and a set of weighted edges $E$, we predict the set of edges forming the minimum spanning tree (MST)..

We formulate this as:

$$\texttt{v}_{i_1}\texttt{v}_{j_1}\texttt{w}_1\ldots\texttt{v}_{i_m}\texttt{v}_{j_m}\texttt{w}_m\texttt{=}\texttt{v}_{p_1}\texttt{v}_{q_1}\ldots\texttt{v}_{p_{n-1}}\texttt{v}_{q_{n-1}}\texttt{<EOS>}$$

Where $(\texttt{v}_{i_k}, \texttt{v}_{j_k}, \texttt{w}_k)$ represents an edge with weight $\texttt{w}_k$, and $\{(\texttt{v}_{p_1}, \texttt{v}_{q_1}), \ldots, (\texttt{v}_{p_{n-1}}, \texttt{v}_{q_{n-1}})\}$ are the edges in the MST.

\textbf{Maximum Subarray}

Given a sequence of $n$ integers $k_1, k_2, \ldots, k_n$ where $k_i \in [-9, 9]$, we predict the contiguous subarray with the maximum sum.

We formulate this as:

$$\texttt{k}_1\texttt{k}_2\ldots\texttt{k}_n\texttt{=}\texttt{k}_i\texttt{k}_{i+1}\ldots\texttt{k}_j\texttt{<EOS>}$$

Where $\texttt{k}_i\texttt{k}_{i+1}\ldots\texttt{k}_j$ is the maximum sum subarray ($i \leq j$), and for a single-element result, only $\texttt{k}_i$ is the output.

\textbf{Activity Selection}

Given a sequence of $n$ activities represented by their start times $(s_1, s_2, \ldots, s_n)$ and finish times $(f_1, f_2, \ldots, f_n)$, we predict the largest subset of non-overlapping activities.

We formulate this as:

$$\texttt{s}_1\texttt{s}_2\ldots\texttt{s}_n\texttt{f}_1\texttt{f}_2\ldots\texttt{f}_n\texttt{=}\texttt{s}_{i_1}\texttt{f}_{i_1}\ldots\texttt{s}_{i_k}\texttt{f}_{i_k}\texttt{<EOS>}$$

Where $\texttt{s}_{i_1}\texttt{f}_{i_1}\ldots\texttt{s}_{i_k}\texttt{f}_{i_k}$ represents the selected non-overlapping activities in ascending order of finish times.

\section{Experimental Methodology}

\subsection{Generating Samples}

\paragraph{Binary Tasks.} For addition and multiplication, pairs of n-bit binary numbers $(a,b)$ are uniformly sampled without replacement. To prevent memorization, if a pair $(a,b)$ appears in the training set, then $(b,a)$ is removed from the validation and test sets. The input for majority of majorities is a single bit string, which is sampled uniformly without replacement.

\paragraph{Graph Tasks.} For graph-based tasks (breadth first search, depth first search, shortest path, minimum spanning tree and topological sorting), we uniformly sample non-isomorphic undirected connected graphs, using the graph dataset of \citet{mckay_graphs}, and randomly permute the vertex labels. For topological sorting, edge directions are determined by randomly sampling a vertex ordering.

\paragraph{Integer Sequence Tasks} For maximum subarray and activity selection, we uniformly sample multisets without replacement.

\subsection{Estimating Scaling Laws (Additional Details)}

During grid search, we filter out hyperparameter combinations that exceed a pre-defined maximum number of steps.\footnote{Binary addition, the first task investigated, did not have this restriction on number of steps.} We ensure at least one trained model across compute budgets reaches 100\% test accuracy on a 1000-example held-out set.

\begin{table}[ht]
\begin{center}
\begin{tabular}{ll}
\toprule
\multicolumn{1}{c}{\bf Task}  &\multicolumn{1}{c}{\bf Input Sizes} \\
\midrule
Addition                & 8, 16, 32, 64, 128 \\
Multiplication          & 16, 32 \\
Majority of Majorities    & 32, 64 \\
Breadth First Search    & 10, 11 \\
Depth First Search      & 10, 11 \\
Shortest Path           & 10, 11 \\
Topological Sorting     & 10, 11 \\
Minimum Spanning Tree   & 10, 11 \\
Maximum Subarray        & 8, 16, 32, 64 \\
Activity Selection       & 8, 16, 32 \\
\bottomrule
\end{tabular}
\end{center}
\caption{Computational tasks and their corresponding input sizes used in our experiments.}\label{tab:tasks}
\end{table}

\begin{table}[h!]
\begin{center}
\begin{tabular}{ll}
\toprule
\multicolumn{1}{c}{\bf Hyperparameter}  &\multicolumn{1}{c}{\bf Range} \\
\midrule
Model Dimension         & [8, 16, 32, 64, 128, 256, 512] \\
Number of Layers        & [4, 16] \\
Number of Heads         & 4 \\
Batch Sizes             & [8, 64] \\
Peak Learning Rate      & [$10^{-1}$, $10^{-2}$, $10^{-3}$, $10^{-4}$] \\
Maximum Steps           & $10^5$ ($10^7$ for compute $> 10^{15}$ FLOPs) \\
\bottomrule
\end{tabular}
\end{center}
\caption{Hyperparameter ranges used in our grid search.}\label{tab:hyperparams}
\end{table}

\begin{table}[ht]
\begin{center}
\begin{tabular}{ll}
\toprule
\multicolumn{1}{c}{\bf Component}  &\multicolumn{1}{c}{\bf Implementation} \\
\midrule
Normalization           & Pre-Norm, RMSNorm \\
Positional Embeddings   & RoPE \\
Feed-forward Network    & SwiGLU \\
AdamW betas             & 0.9, 0.95 \\
Linear Bias             & False \\
Learning Rate Scheduler & Linear Warmup \\ & (from 0.01 of peak LR \\ & over 10\% of training steps) \\ & + Cosine Decay to 0.1 of peak LR \\
\bottomrule
\end{tabular}
\end{center}
\caption{Architectural modifications used in our Transformer++ implementation.}\label{tab:transformer}
\end{table}

\subsection{Training Feature Probes}
\label{subsec:training-feature-probes}

\subsubsection{Transformer Architecture and Residual Stream}

Consider a transformer model with $L$ layers. For each layer $l \in \{1, 2, ..., L\}$ and token position $t$, we define the layer computation as:

\begin{align}
\mathbf{x}_{\text{mid}}^{(l,t)} &= \mathbf{x}_{\text{pre}}^{(l,t)} + \sum_{\text{head } h} \text{attn}^{(l,h)}\left(\mathbf{x}_{\text{pre}}^{(1,t)}, \mathbf{x}_{\text{pre}}^{(1,1:t)}\right) \label{eq:mid-residual}\\
\mathbf{x}_{\text{post}}^{(l,t)} &= \mathbf{x}_{\text{mid}}^{(l,t)} + \text{MLP}^{(l)}\left(\mathbf{x}_{\text{mid}}^{(l,t)}\right) \label{eq:post-residual}
\end{align}

where:
\begin{itemize}
    \item $\mathbf{x}_{\text{pre}}^{(l,t)} \in \mathbb{R}^d$ is  input to the layer $l$ at position $t$ (the pre-residual stream). $d$ is the transformer model dimension.
    \item $\mathbf{x}_{\text{mid}}^{(l,t)} \in \mathbb{R}^d$ is the mid-residual stream (after attention)
    \item $\mathbf{x}_{\text{post}}^{(l,t)} \in \mathbb{R}^d$ is the output of layer $l$ (post-residual stream)
    \item $\text{attn}^{(l,h)}$ denotes the $h$-th attention head in layer $l$
    \item $\text{MLP}^{(l)}$ denotes the feedforward network in layer $l$
\end{itemize}

We train linear probes on the output of the layer, $\mathbf{x}_{\text{post}}^{(l,t)}$ for each layer $l$ and token position $t$.

\subsubsection{Probe Training Procedure}

For each feature $f$ at token position $t$ and layer $l$, we train a probe $p_{f,l,t}$ on the output of layer $l$, $\mathbf{x}_{\text{post}}^{(l,t)}$. The type of probe depends on the feature:

\paragraph{Binary Features.} For binary feature $f \in \{0, 1\}$, we train a logistic regression classifier:

\begin{equation}
p_{f,l,t}(\mathbf{x}_{post}^{(l, t)}) = \sigma(\mathbf{w}_{f,l,t}^T \mathbf{x}_{post}^{(l, t)} + b_{f,l,t})
\end{equation}
where $\sigma$ is the sigmoid function, $\mathbf{w}_{f,l,t} \in \mathbb{R}^d$, and $b_{f,l,t} \in \mathbb{R}$. The following features are binary: \textit{first\_operand}, \textit{carry}, \textit{is\_prev\_negative}.

\paragraph{Multi-valued Features.} Features \textit{queue} \& \textit{adjacency\_list} represent list of binary variables. For example, \textit{adjacency\_list} at token $t$ is a list $(e_1, \dots, e_k)$ where $e_j \in {0,1}$ represents whether vertex $v_t$ is connected with vertex $v_j$. To detect such features, we train $k$ independent logistic classifiers:
\begin{equation}
p_{f,l,t}^{(i)}(\mathbf{x}_{post}^{(l, t)}) = \sigma(\mathbf{w}_{f,l,t}^{(i)T} \mathbf{x}_{post}^{(l, t)} + b_{f,l,t}^{(i)}) \quad \text{for } i = 1, ..., k
\end{equation}

\paragraph{Real-valued Features.} For continuous features, \textit{max\_ending\_here} \& \textit{retrieve\_start\_times}, we train a linear regressor:
\begin{equation}
p_{f,l,t}(\mathbf{x}) = \mathbf{w}_{f,l,t}^T \mathbf{x} + b_{f,l,t}
\end{equation}

\subsubsection{Training Configuration}

All probes are trained using the configuration noted in Table \ref{tab:probe-hyperparameters}. 

\begin{table}[h]
\centering
\begin{tabular}{ll}
\toprule
\textbf{Parameter} & \textbf{Value} \\
\midrule
Training examples & 10,000 \\
Regularization strength ($C$) & 100 \\
Fit intercept & True \\
Maximum iterations & 1,000 \\
Optimizer & L-BFGS (scikit-learn default) \\
\bottomrule
\end{tabular}
\caption{Probe training hyperparameters}
\label{tab:probe-hyperparameters}
\end{table}

\subsubsection{Probe Selection}

Given a trained model with compute budget $B$, we select the best probe for each feature $f$ and token position $t$ as follows:
\begin{equation}
l^*_{f,t} = \argmin_{l \in \{1, ..., L\}} \mathcal{L}_{\text{train}}(p_{f,l,t})
\end{equation}
where $\mathcal{L}_{\text{train}}$ denotes the training loss (cross-entropy for classification, mean squared error for regression). The test performance is then evaluated using probe $p_{f,l^*_{f,t},t}$ on a held-out test set of 1,000 examples.

All our training was done on an 8U HGX server with Dual Intel Sapphire Rapids and 8 NVIDIA H100 GPUs. Test accuracies for feature ablation were computed on a machine with Intel(R) Xeon(R) Gold 6230 CPU and NVIDIA GeForce RTX 2080 Ti.

\begin{table*}[ht]
\centering
\footnotesize
\begin{tabular}{ll*{6}{c}}
\toprule
\textbf{Task} & \textbf{Feature} & \multicolumn{2}{c}{\textbf{Baseline}} & \multicolumn{2}{c}{\textbf{Feature Ablation}} & \multicolumn{2}{c}{\textbf{Random Ablation}} \\
\cmidrule(lr){3-4} \cmidrule(lr){5-6} \cmidrule(lr){7-8}
 &  & Acc. (\%) & Loss & Acc. (\%) & Loss & Acc. (\%) & Loss \\
\midrule
Addition (16) & \textit{carry} & 100 & 7.91e-10 & 58.8 & 3.12e-2 & 100 & 7.96e-10 \\ 

 &  & \scriptsize{[99.6, 100]} & \scriptsize{[6.59e-10, 9.29e-10]} & \scriptsize{[55.8, 61.9]} & \scriptsize{[3.06e-02, 3.18e-02]} & \scriptsize{[99.6, 100]} & \scriptsize{[7.73e-10, 8.20e-10]} \\[0.5ex]
 
Addition (32) & \textit{carry} & 100 & 1.53e-10 & 49.6 & 2.87e-2 & 100 & 1.66e-10 \\

 &  & \scriptsize{[99.6, 100]} & \scriptsize{[9.20e-11, 2.42e-10]} & \scriptsize{[46.5, 52.7]} & \scriptsize{[2.70e-2, 3.04e-2]} & \scriptsize{[99.6, 100]} & \scriptsize{[1.497e-10, 1.822e-10]} \\[0.5ex]
 
Addition (64) & \textit{carry} & 100 & 4.68e-10 & 24.9 & 4.60e-2 & 100 & 1.25e-9 \\

 &  & \scriptsize{[99.6, 100]} & \scriptsize{[3.54e-10, 6.16e-10]} & \scriptsize{[22.3, 27.6]} & \scriptsize{[4.39e-2, 4.81e-2]} & \scriptsize{[99.6, 100]} & \scriptsize{[1.20e-9, 1.32e-9]} \\[0.5ex]
 
Addition (16) & \textit{first\_operand} & 100 & 7.91e-10 & 100 & 7.91e-10 & 100 & 8.68e-10 \\

 &  & \scriptsize{[99.6, 100]} & \scriptsize{[6.59e-10, 9.22e-10]} & \scriptsize{[99.62, 100]} & \scriptsize{[6.65e-10, 9.29e-10]} & \scriptsize{[99.6, 100]} & \scriptsize{[8.44e-10, 8.93e-10]} \\[0.5ex]
 
Addition (32) & \textit{first\_operand} & 100 & 1.53e-10 & 7.00 & 7.57e-1 & 99.7 & 3.46e-4 \\

 &  & \scriptsize{[99.6, 100]} & \scriptsize{[8.86e-11, 2.42e-10]} & \scriptsize{[5.40, 8.60]} & \scriptsize{[7.25e-1, 7.89e-1]} & \scriptsize{[99.6, 100]} & \scriptsize{[2.58e-4, 4.43e-4]} \\[0.5ex]
 
Addition (64) & \textit{first\_operand} & 100 & 4.68e-10 & 93.6 & 3.65e-3 & 100 & 1.07e-8 \\

 &  & \scriptsize{[99.6, 100]} & \scriptsize{[3.51e-10, 6.10e-10]} & \scriptsize{[92.1, 95.1]} & \scriptsize{[2.72e-3, 4.66e-3]} & \scriptsize{[99.6, 100]} & \scriptsize{[6.88e-9, 1.57e-8]} \\[0.5ex]
 
Multiplication (16) & \textit{carry} & 76.8 & 1.07e-2 & 56.4 & 5.55e-2 & 76.7 & 1.10e-2 \\ 

 &  & \scriptsize{[74.2, 79.3]} & \scriptsize{[9.79e-3, 1.16e-2]} & \scriptsize{[53.3, 59.5]} & \scriptsize{[4.98e-2, 6.17e-2]} & \scriptsize{[76.2, 77.2]} & \scriptsize{[1.08e-2, 1.17e-2]} \\[0.5ex]
 
Multiplication (16) & \textit{first\_operand} & 76.8 & 1.07e-2 & 76.8 & 1.07e-2 & 76.9 & 1.07e-2 \\ 

 &  & \scriptsize{[74.2, 79.3]} & \scriptsize{[9.79e-3, 1.16e-2]} & \scriptsize{[74.2, 79.3]} & \scriptsize{[9.79e-3, 1.16e-2]} & \scriptsize{[76.2, 77.3]} & \scriptsize{[1.06e-2, 1.09e-2]} \\[0.5ex]
 
Maximum Subarray (64) & \textit{is\_prev\_negative} & 95.6 & 1.51e-2 & 89.9 & 3.15e-2 & 94.4 & 1.84e-2 \\ 

 &  & \scriptsize{[94.3, 96.8]} & \scriptsize{[9.25e-3, 2.19e-2]} & \scriptsize{[88.0, 91.7]} & \scriptsize{[2.32e-2, 4.07e-2]} & \scriptsize{[93.8, 94.3]} & \scriptsize{[1.72e-2, 1.97e-2]} \\[0.5ex]
 
Breadth first search (11) & \textit{queue} & 99.7 & 8.71e-4 & 54.6 & 1.30e-1 & 98.2 & 4.87e-3 \\

 &  & \scriptsize{[99.1, 99.9]} & \scriptsize{[3.23e-4, 1.63e-3]} & \scriptsize{[51.5, 57.6]} & \scriptsize{[1.20e-1, 1.40e-1]} & \scriptsize{[98.1, 98.4]} & \scriptsize{[4.44e-3, 5.35e-3]} \\[0.5ex]

 &  & & & & & & \\
 
\bottomrule
\end{tabular}
\caption{Test accuracy \& loss with and without feature ablations on 1000 examples for different tasks and input sizes. Baseline refers to the language model's accuracy without any perturbations. For random ablation, we report the mean over 32 trials. 95\% confidence intervals shown in smaller text below each value.}
\label{tab:all_ablation_metrics}
\end{table*}

\clearpage

\begin{figure*}[ht]
\begin{center}
\includegraphics[width=\linewidth]{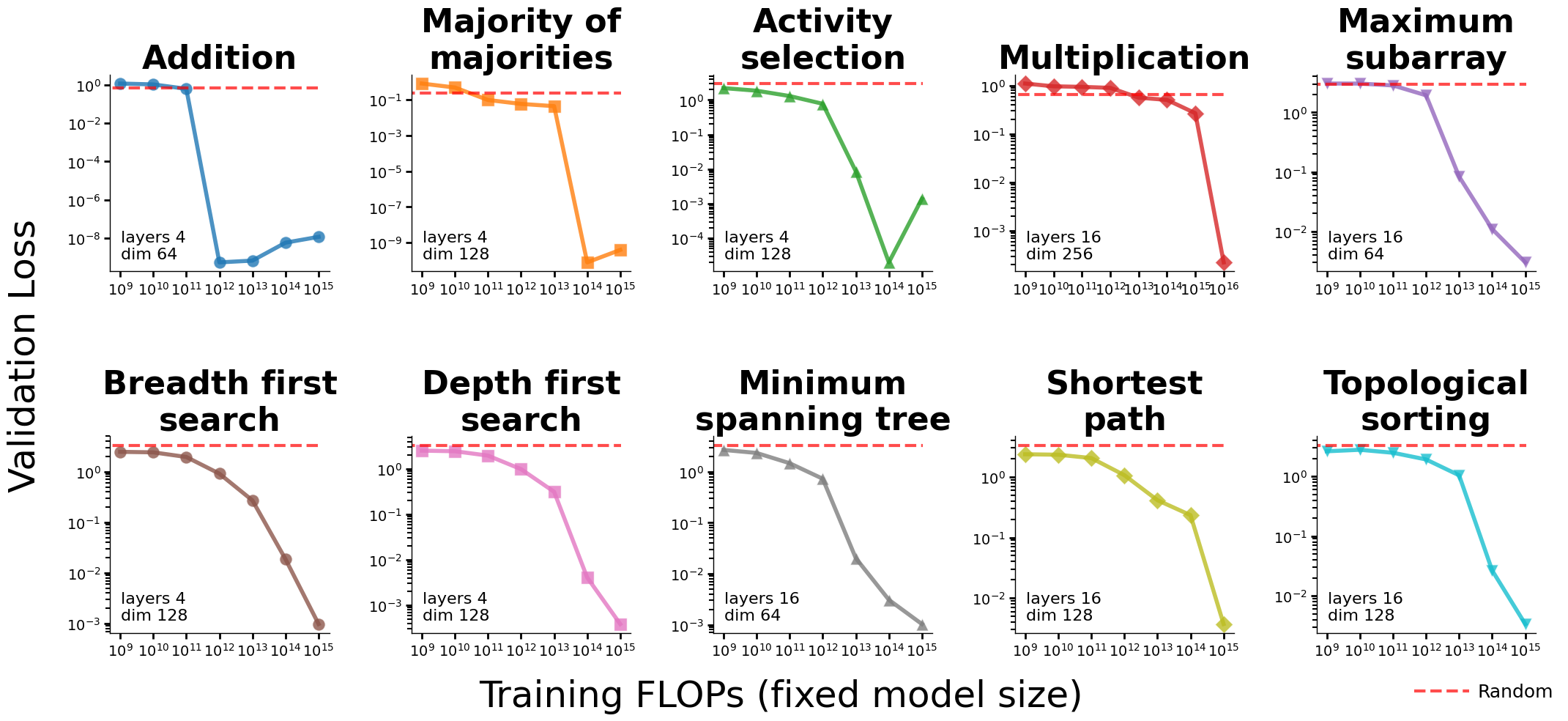}
\end{center}
\caption{Despite holding model size constant, model performance shows abrupt improvement across various amount of training compute. Plot the minimum validation Loss for each compute. Model sizes are compute-optimal for the earliest training compute where test accuracy is 100\%. The input sizes are the same as in Figure \ref{fig:val-loss-across-tasks}.}
\label{fig:val-loss-fixed-model-size-across-tasks}
\end{figure*}

\begin{figure*}[ht]
\begin{center}
\includegraphics[width=\linewidth]{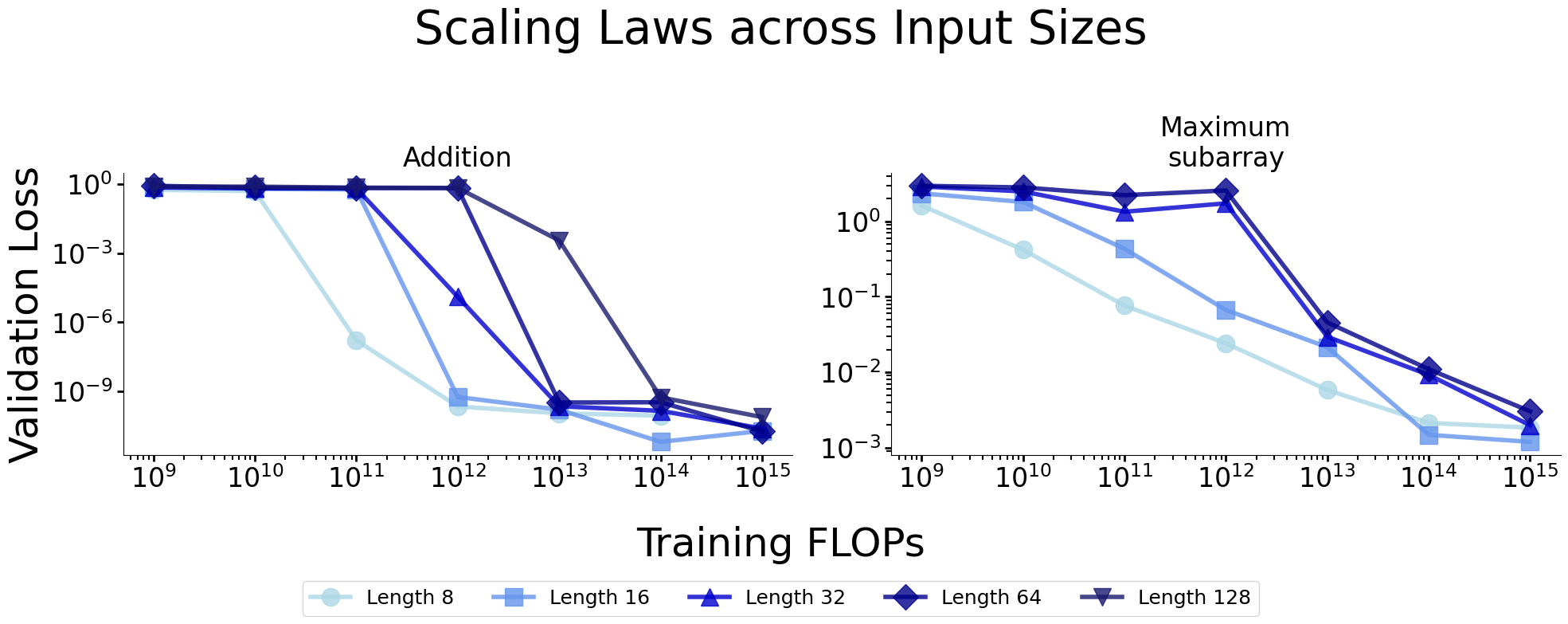}
\end{center}
\caption{Validation loss for best run from grid search vs  training FLOPs  for addition \& maximum subarray across different input sizes}
\label{fig:val-loss-add-maxsubarr-all-lengths}
\end{figure*}

\begin{figure*}[ht]
\begin{center}
\includegraphics[width=\linewidth]{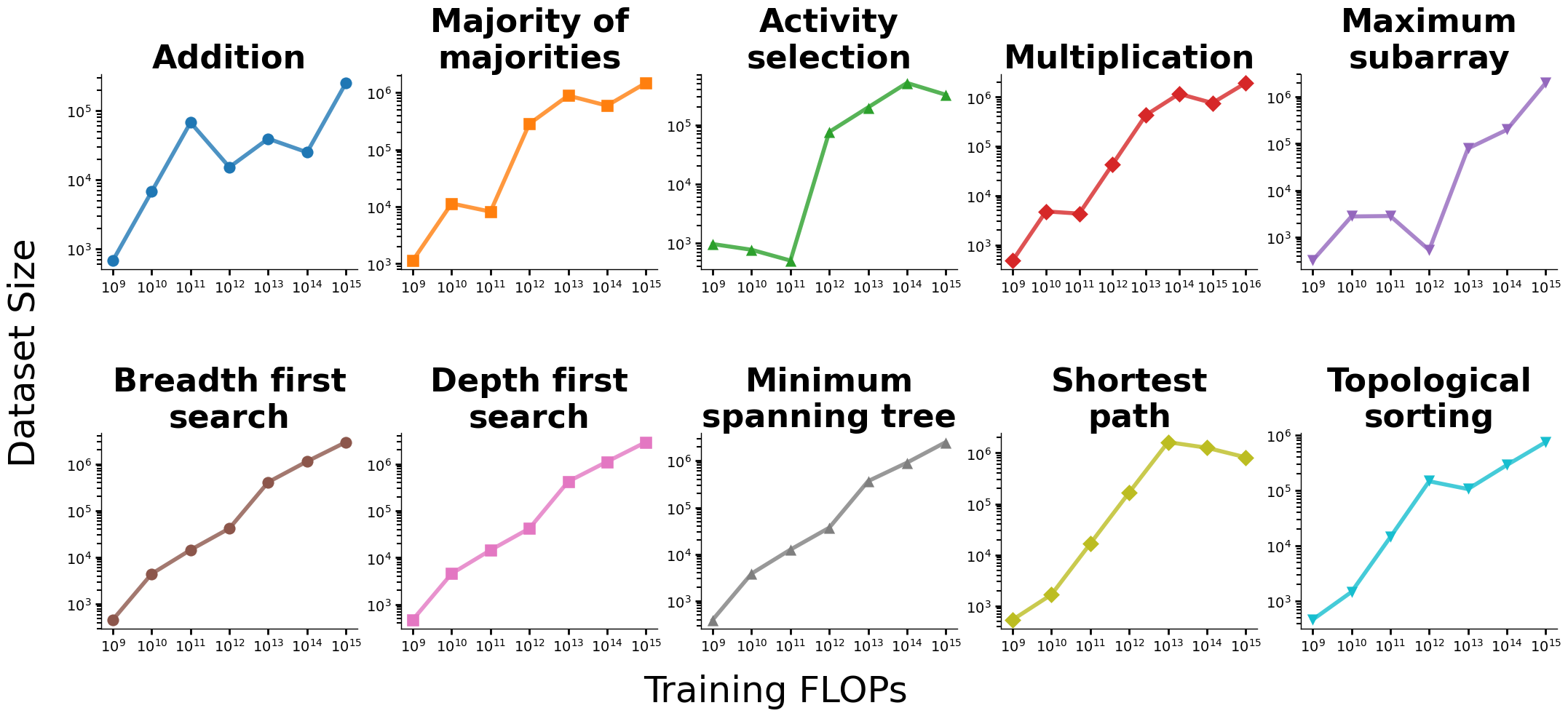}
\end{center}
\caption{Compute-optimal dataset size (\# of training examples) vs training FLOPs}
\label{fig:val-loss-add-maxsubarr-all-lengths}
\end{figure*}

\begin{figure*}[ht]
\begin{center}
\includegraphics[width=\linewidth]{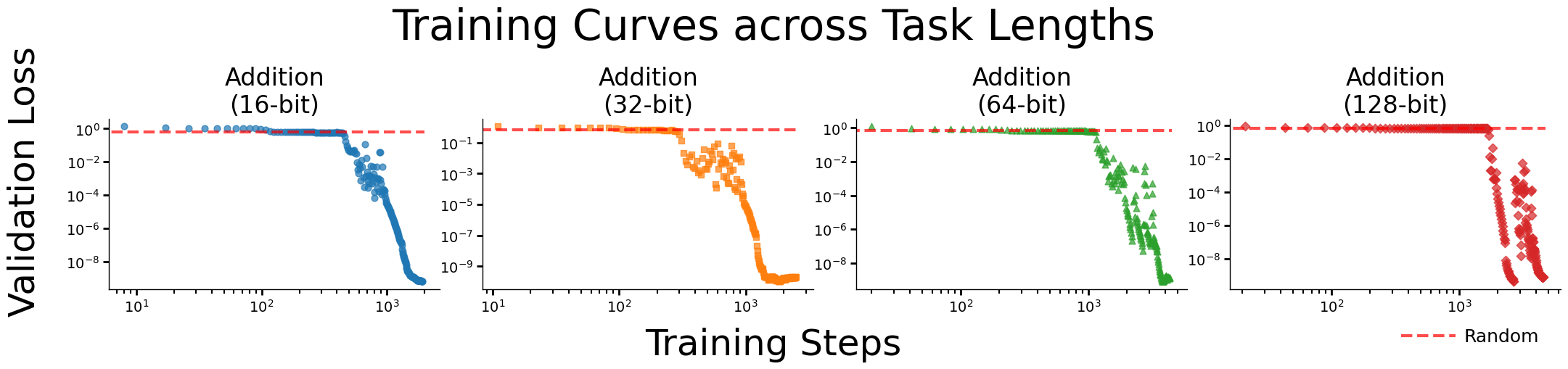}
\end{center}
\caption{Validation loss vs training steps for a single training run for addition across input sizes. Compute-optimal training run is selected for the earliest training FLOPs in the same fashion Figure \ref{fig:val-loss-training-steps-multiple-tasks}}
\label{fig:val-loss-addition-training-all-lengths}
\end{figure*}
\end{document}